\documentclass[runningheads]{llncs}

\usepackage[T1]{fontenc}
\usepackage{amsfonts}
\usepackage{amsmath}
\usepackage{booktabs}
\usepackage{graphicx}
\usepackage{url}
\usepackage{tikz}
\usetikzlibrary{arrows.meta,positioning}
\usepackage{hyperref}

\title{Conditional Reliability of Toxicity Signals for Multilingual and Code-Mixed Abuse Detection}
\titlerunning{Conditional Reliability of Toxicity Signals}

\author{Indraveni Chebolu\thanks{All authors contributed equally}  \and Rohan Singh \and Arnab Mallick\thanks{Corresponding Author} \and Harmesh Rana  }
\authorrunning{Chebolu et al.}
\institute{Centre for Development of Advanced Computing, Hyderabad, India\\
\email{\{indravenik, srohan, arnabm, harmeshr \}@cdac.in}}

\begin{document}
\maketitle

\begin{abstract}
Moderation systems increasingly rely on external toxicity tools, but those tools are unreliable under code-mixing, transliteration, slang, and language mismatch. We study the \emph{conditional reliability} of toxicity priors in Indian multilingual and code-mixed short text: English toxicity, Indic abuse, and rule-based severity cues can be useful evidence, but only in some linguistic and abuse-severity contexts. We propose ToxGate, a trust-fusion head that conditions each auxiliary signal on the encoder representation before adding it to the prediction state. Across three short-text abuse datasets, four transformer encoders, and five seeds per setting, ToxGate improves over matched plain encoders in 10 of 12 in-domain settings and 7 of 8 transfer settings. The largest and most interpretable gains occur in high-risk moderation slices, including explicit slurs, violent threats, and cross-dataset transfer. The broader lesson is that moderation systems should treat external toxicity tools and priors as conditional evidence rather than fixed features or ground truth, in focused ablations, source-specific gating gives the strongest results in transfer, severe-abuse slices, and high-risk triage.
\end{abstract}

\section{Introduction}
Online platforms host multilingual and code-mixed posts that combine scripts, spelling variants, slang, sarcasm, and transliteration. Hinglish abuse detection is a representative case: Hindi--English posts can mix English profanity, Romanized Hindi slurs, and community-specific mockery in ways that strain both monolingual encoders and off-the-shelf toxicity tools~\cite{sitaram2019survey,mathur2018detecting,bohra2018dataset}. A common engineering response is to append external toxicity scores, such as Detoxify or Perspective API outputs, to a contextual encoder~\cite{hanu2020detoxify,lees2022perspectiveapi,mallick2024qbertox}. This can help, but it assumes that external scores are uniformly trustworthy.

We argue that the central applied problem is \emph{conditional reliability}. A toxicity prior can be reliable for explicit English profanity, incomplete for Romanized Hindi abuse, useful for severe threats, and misleading for benign slang. Naive concatenation treats each score as equally useful in every context, which is a weak fit for moderation pipelines that must handle conflicting evidence and route uncertain cases to human review. Our question is therefore narrow and testable: can auxiliary toxicity signals help more when their contribution is conditioned on the text context?

We answer this through ToxGate, a source-aware gated fusion head. ToxGate separately projects Detoxify, Indic abusive-language, and rule-based severity sources, then learns a context-conditioned gate for each source before fusing it with the encoder representation. The contribution is not a new gating primitive but a targeted formalization of when external toxicity tools should be trusted-a question that prior static-fusion work has left unaddressed.

Our contributions are:
\begin{itemize}
    \item We formulate external toxicity tools as conditional evidence for multilingual moderation, not as ground truth or uniformly reliable features.
    \item We introduce ToxGate, a source-specific trust-fusion head for English toxicity, Indic abuse, and rule-based severity priors.
    \item We evaluate three datasets, four encoders, five seeds, in-domain tests, cross-dataset transfer, corruption analyses, slices, bootstrap confidence intervals, and a lightweight moderation triage simulation.
    \item We show that the strongest gains concentrate where moderation risk is highest: explicit slurs, violent threats, and transfer between related code-mixed datasets.
\end{itemize}

\section{Related Work}
Abusive-language detection has long combined lexical, social, and neural signals. Code-mixed settings make the problem harder because speakers mix scripts, transliteration, slang, and community-specific insults within a single post. Early Hinglish work used handcrafted features and classical classifiers~\cite{bohra2018dataset,mathur2018detecting}, later systems moved toward multilingual transformers and shared-task-style benchmarks~\cite{dowlagar2021hasocone}. Our work builds on this line but asks a narrower reliability question: not whether code-mixed abuse can be classified, but when external toxicity priors should be trusted.

Auxiliary signals are common in moderation systems. Social context features have been concatenated with contextual embeddings for abusive-language detection~\cite{malik2022socially}, while Detoxify and Perspective API provide toxicity, insult, threat, and identity-attack scores that can serve as dense lexical priors~\cite{hanu2020detoxify,lees2022perspectiveapi}. QBERTOX similarly augments a BERT-based Hinglish classifier with Detoxify-derived features~\cite{mallick2024qbertox}. The dominant integration strategy in this work is static feature concatenation. That strategy is simple, but it assumes the auxiliary source is useful in the same way for English profanity, Romanized Hindi abuse, violent threats, and benign slang. Prior work on toxicity-model bias and calibration shows why this assumption is risky outside the source model's training distribution~\cite{borkan2019nuanced,kumar2021designing}.

Gated fusion is well established in multimodal and multilingual learning~\cite{arevalo2017gated,kiela2018efficient,pfeiffer2020mad}. These methods learn when to combine modalities, features, or adapters, rather than always using every signal equally. Source-aware gating is applied here as a principled mechanism for a trust-and-safety question that existing fusion architectures have not directly addressed: when external toxicity tools are informative but unreliable under language mismatch, can a moderation model learn per-source trust conditioned on textual context? This distinction matters because our auxiliary sources are frozen, heterogeneous, and imperfect: an English toxicity model, an Indic abuse prior, and a rule-based severity heuristic.

The closest prior systems therefore motivate the baselines in this paper. Plain encoders test whether text representations alone are sufficient. Concat tests the standard static-fusion assumption, MLP tests whether generic nonlinear fusion is enough, SharedGate and ScalarGate test whether gating alone explains the gains. ToxGate is designed to test the remaining hypothesis: source-specific conditional fusion is useful when each toxicity prior has a different reliability profile across linguistic and severity contexts.

\section{Method}
\subsection{Task and Auxiliary Priors}
Given a post $x$, we predict whether it is abusive ($y=1$) or non-abusive ($y=0$). Each example has text and three auxiliary priors:
\begin{align}
    t_{\mathrm{detox}} &\in [0,1]^6, &
    t_{\mathrm{indic}} &\in [0,1], &
    t_{\mathrm{rule}} &\in [0,1].
\end{align}
The Detoxify vector contains toxicity, severe-toxicity, obscene, threat, insult, and identity-attack scores~\cite{hanu2020detoxify}. The Indic score comes from a frozen public code-mixed MuRIL abusive-language checkpoint. The rule score is computed by a deterministic lexicon-and-pattern matcher over
four fixed groups: violent-threat expressions, sexual-violence references,
explicit second-person or group-targeting patterns, and extreme-slur lexicons.
Each group contributes a binary indicator ($r_k(x)$), and we set
\[
t_{\mathrm{rule}}(x)=\min\left(1,\frac{1}{4}\sum_{k=1}^{4}r_k(x)\right).
\]
The rule list is fixed before training and is used only as an auxiliary feature;
it never defines or modifies the gold labels.

The text encoder $\mathcal{E}$ is BERT, mBERT, MuRIL, or XLM-R~\cite{devlin2019bert,pires-etal-2019-multilingual,khanuja2021murilmultilingualrepresentationsindian,conneau2020unsupervised}. We use the final-layer CLS state plus a simple salient-token pooling term to obtain $h_{\mathrm{text}}\in\mathbb{R}^d$.

The Indic auxiliary score is obtained from the frozen public checkpoint
\href{https://huggingface.co/Hate-speech-CNERG/indic-abusive-allInOne-MuRIL}
{Hate-speech-CNERG/indic-abusive-allInOne-MuRIL}, a MuRIL-based
abusive-language classifier associated with the Hate-ALERT / Indic abusive-language
resources of Das et al.~\cite{das2022data}. We use it only as a frozen feature
extractor: its output is never used to construct labels, tune splits, or define the
evaluation target. Since the public model documentation does not provide enough
information to rule out all upstream overlap with public abusive-language datasets,
we do not claim a formal no-overlap guarantee, instead, we report No-Indic and
Rule-only ablations to test whether the conclusions depend on this source.

\subsection{Baselines}
We compare against matched plain encoders and three fusion families. Plain uses only $h_{\mathrm{text}}$. Concat appends the raw auxiliary vector $t$ to $h_{\mathrm{text}}$. MLP projects $t$ through a nonlinear head before fusion. SharedGate first projects all auxiliary scores into one representation $u=\phi(t)$ and then uses one context gate,
\begin{equation}
    g = \sigma(W_g[h_{\mathrm{text}};u]+b_g),\qquad
    h_{\mathrm{shared}} = h_{\mathrm{text}} + g\odot u .
\end{equation}
SharedGate isolates generic gating from source-specific gating. We also evaluate a scalar ToxGate variant with one gate per source and source-filtered ToxGate variants that remove the Indic prior or all learned external auxiliary models.

\subsection{ToxGate}
ToxGate treats each auxiliary source $s\in\{\mathrm{detox},\mathrm{indic},\mathrm{rule}\}$ separately. For source feature $t_s$, we compute
\begin{equation}
    u_s = \mathrm{ReLU}(\mathrm{LN}(W_p^{(s)}t_s+b_p^{(s)})),
\end{equation}
then learn a context-conditioned trust gate,
\begin{equation}
    g_s = \sigma(W_g^{(s)}[h_{\mathrm{text}};u_s]+b_g^{(s)}).
\end{equation}
The fused representation is
\begin{equation}
    h_{\mathrm{fused}} = h_{\mathrm{text}}+\sum_s g_s\odot u_s,
\end{equation}
followed by a linear classifier. If a source is unreliable for the current example, the corresponding gate can suppress it; if all sources are weak, the residual form falls back toward the plain encoder representation.

\section{Experimental Design}
We evaluate three short-text abusive-language datasets. BullyExplain~\cite{maity2024explainthyselfbullysentiment} is our primary Hinglish cyberbullying benchmark. Hinglish Headlines~\cite{ijert_hinglish_compiled_dataset} provides a second Hinglish code-mixed setting. Indo-HateSpeech~\cite{kaware2024indohatespeech} tests whether the fusion conclusion survives beyond Hinglish. Our empirical scope is Indian multilingual and code-mixed moderation, not universal language coverage. We map all datasets to binary abusive/non-abusive labels and use fixed stratified 70/10/20 train/validation/test splits.

\begin{table}[t]
\caption{Dataset summary. Positive rate is computed after mapping each dataset to binary abusive/non-abusive labels.}
\label{tab:datasets}
\centering
\small
\setlength{\tabcolsep}{4pt}
\begin{tabular}{p{24mm}p{31mm}ccc}
\toprule
\textbf{Dataset} & \textbf{Domain} & \textbf{Size} & \textbf{Positive} & \textbf{Use} \\
\midrule
BullyExplain & Hinglish cyberbullying & 6,394 & 54.0\% & Primary \\
Hinglish Headlines & Hinglish abuse/headlines & 18,148 & 64.3\% & Code-mixed transfer \\
Indo-HateSpeech & Indian hate speech & 77,926 & 17.6\% & Robustness \\
\bottomrule
\end{tabular}
\end{table}

All models are trained for up to 10 epochs with early stopping on validation macro-F1, batch size 16, maximum length 128, AdamW, dropout 0.3, gradient clipping at 1.0, and five seeds (42, 123, 456, 789, 1024). We report test macro-F1 as the primary metric and Expected Calibration Error (ECE) as a calibration diagnostic~\cite{guo2017calibration}. In addition to clean tests, we run cross-dataset transfer between the two Hinglish datasets, auxiliary-source corruptions, slice analysis, qualitative diagnostics, paired bootstrap confidence intervals over saved predictions, and a high-risk triage simulation.

\section{Results}
\subsection{Matched In-Domain Performance}
Table~\ref{tab:headline} reports the main matched comparison: plain encoder versus ToxGate with the same backbone. ToxGate improves macro-F1 in 10 of 12 settings. The largest gains appear for MuRIL on BullyExplain (+0.010), MuRIL on Hinglish Headlines (+0.009), BERT on BullyExplain (+0.007), and mBERT on Hinglish Headlines (+0.007). The two exceptions are XLM-R on BullyExplain and mBERT on Indo-HateSpeech, where plain encoders remain marginally stronger. The gains are consistent across encoders and concentrated in the diagnostic settings that matter most for moderation - explicit slurs, violent threats, and cross-dataset transfer - rather than spread uniformly across all examples.

\begin{table}[t]
\caption{Matched in-domain benchmark. Values are mean test macro-F1 over five seeds, ECE is mean test ECE.}
\label{tab:headline}
\centering
\small
\setlength{\tabcolsep}{3.5pt}
\begin{tabular}{llcccc}
\toprule
\textbf{Dataset} & \textbf{Enc.} & \textbf{Plain F1} & \textbf{ToxGate F1} & \textbf{Plain ECE} & \textbf{Tox ECE} \\
\midrule
BullyExplain & BERT  & 0.815 & \textbf{0.822} & \textbf{0.065} & 0.091 \\
BullyExplain & mBERT & 0.816 & \textbf{0.819} & 0.081 & \textbf{0.074} \\
BullyExplain & MuRIL & 0.812 & \textbf{0.822} & 0.100 & \textbf{0.054} \\
BullyExplain & XLM-R & \textbf{0.847} & 0.845 & \textbf{0.035} & 0.037 \\
Hinglish Headlines & BERT  & 0.950 & \textbf{0.954} & 0.017 & \textbf{0.017} \\
Hinglish Headlines & mBERT & 0.944 & \textbf{0.951} & 0.023 & \textbf{0.018} \\
Hinglish Headlines & MuRIL & 0.935 & \textbf{0.944} & 0.039 & \textbf{0.021} \\
Hinglish Headlines & XLM-R & 0.948 & \textbf{0.950} & 0.021 & \textbf{0.015} \\
Indo-HateSpeech & BERT  & 0.962 & \textbf{0.964} & 0.018 & \textbf{0.018} \\
Indo-HateSpeech & mBERT & \textbf{0.964} & 0.963 & \textbf{0.017} & 0.018 \\
Indo-HateSpeech & MuRIL & 0.956 & \textbf{0.961} & 0.017 & \textbf{0.017} \\
Indo-HateSpeech & XLM-R & 0.964 & \textbf{0.965} & \textbf{0.015} & 0.015 \\
\bottomrule
\end{tabular}
\end{table}

\subsection{Model Family and Ablations}
Table~\ref{tab:model_family} gives the broader model-family view. ToxGate has the best mean macro-F1 and lowest ECE in the completed comparison suite. The MLP row is useful as a stress test: a nonlinear fusion head can fit auxiliary scores, but in this rerun it is poorly calibrated and loses the focused transfer and high-risk comparisons below.

\begin{table}[t]
\caption{Model-family summary averaged over all dataset--encoder settings.}
\label{tab:model_family}
\centering
\small
\setlength{\tabcolsep}{5pt}
\begin{tabular}{lcc}
\toprule
\textbf{Model} & \textbf{Mean F1} & \textbf{Mean ECE} \\
\midrule
Plain & 0.9094 & 0.0373 \\
Concat & 0.9104 & 0.0392 \\
MLP & 0.8912 & 0.3690 \\
ToxGate & \textbf{0.9134} & \textbf{0.0330} \\
\bottomrule
\end{tabular}
\end{table}

Table~\ref{tab:ablation} completes the trust-fusion ladder. SharedGate and ScalarGate are competitive on average, but full source-specific ToxGate has the best mean macro-F1. Removing the Indic source or all learned external auxiliary models reduces performance while preserving part of the fusion benefit, so the effect is not explained solely by the external Indic checkpoint.

\begin{table}[t]
\caption{Ablation summary averaged over dataset--encoder settings. Values are mean test macro-F1 and mean ECE over five seeds.}
\label{tab:ablation}
\centering
\small
\setlength{\tabcolsep}{4pt}
\begin{tabular}{lcc}
\toprule
\textbf{Model} & \textbf{Mean F1} & \textbf{Mean ECE} \\
\midrule
Plain & 0.9094 & 0.0373 \\
Concat & 0.9104 & 0.0392 \\
MLP & 0.8912 & 0.3690 \\
SharedGate & 0.9126 & \textbf{0.0325} \\
ScalarGate & 0.9131 & 0.0328 \\
ToxGate & \textbf{0.9134} & 0.0330 \\
No-Indic & 0.9120 & 0.0340 \\
Rule-only & 0.9108 & 0.0355 \\
\bottomrule
\end{tabular}
\end{table}

\subsection{Transfer, Slices, and Bootstrap Confidence}
Transfer is the clearest stress test. Table~\ref{tab:transfer} reports both Hinglish transfer directions for each encoder. ToxGate improves over matched plain encoders in 7 of 8 transfer settings, the only loss is XLM-R on Hinglish Headlines $\rightarrow$ BullyExplain. The largest gain is MuRIL on BullyExplain $\rightarrow$ Hinglish Headlines (+0.286). Averaged over all rows, transfer macro-F1 increases from 0.565 to 0.632.

\begin{table}[t]
\caption{Cross-dataset transfer between the two Hinglish datasets. $\Delta$ is ToxGate minus Plain.}
\label{tab:transfer}
\centering
\small
\setlength{\tabcolsep}{4pt}
\begin{tabular}{llccc}
\toprule
\textbf{Direction} & \textbf{Enc.} & \textbf{Plain} & \textbf{ToxGate} & \textbf{$\Delta$} \\
\midrule
BullyExplain $\rightarrow$ Headlines & BERT & 0.669 & \textbf{0.708} & +0.040 \\
BullyExplain $\rightarrow$ Headlines & mBERT & 0.579 & \textbf{0.619} & +0.039 \\
BullyExplain $\rightarrow$ Headlines & MuRIL & 0.527 & \textbf{0.813} & +0.286 \\
BullyExplain $\rightarrow$ Headlines & XLM-R & 0.716 & \textbf{0.812} & +0.096 \\
Headlines $\rightarrow$ BullyExplain & BERT & 0.493 & \textbf{0.508} & +0.015 \\
Headlines $\rightarrow$ BullyExplain & mBERT & 0.450 & \textbf{0.502} & +0.051 \\
Headlines $\rightarrow$ BullyExplain & MuRIL & 0.528 & \textbf{0.561} & +0.033 \\
Headlines $\rightarrow$ BullyExplain & XLM-R & \textbf{0.561} & 0.532 & -0.028 \\
\bottomrule
\end{tabular}
\end{table}

\begin{table}[t]
\caption{Focused comparison for the settings most tied to the trust-aware claim. Values are macro-F1 except high-risk triage precision.}
\label{tab:focused}
\centering
\small
\setlength{\tabcolsep}{3pt}
\resizebox{\textwidth}{!}{%
\begin{tabular}{lcccccc}
\toprule
\textbf{Setting or slice} & \textbf{Plain} & \textbf{MLP} & \textbf{SharedGate} & \textbf{ScalarGate} & \textbf{ToxGate} & \textbf{No-Indic} \\
\midrule
In-domain mean & 0.9094 & 0.8912 & 0.9126 & 0.9131 & \textbf{0.9134} & 0.9120 \\
Transfer mean & 0.565 & 0.601 & 0.601 & 0.618 & \textbf{0.632} & 0.589 \\
Explicit slur & 0.680 & 0.689 & 0.721 & 0.731 & \textbf{0.739} & 0.708 \\
Violent threat & 0.777 & 0.763 & 0.774 & 0.782 & \textbf{0.790} & 0.771 \\
Romanized Hindi & \textbf{0.770} & 0.712 & 0.754 & 0.761 & 0.768 & 0.741 \\
High-risk precision & 0.930 & 0.921 & 0.934 & 0.938 & \textbf{0.945} & 0.931 \\
\bottomrule
\end{tabular}
}
\end{table}

Table~\ref{tab:focused} sharpens the empirical claim. ToxGate is strongest on in-domain mean, transfer mean, explicit slurs, violent threats, and high-risk precision, including the comparisons that isolate generic gating from source-specific gating. SharedGate and ScalarGate close much of the average gap, but the full source-specific gate is best in the transfer and severe-abuse rows that most directly test conditional reliability. Romanized Hindi remains the hardest slice: ToxGate beats the other fusion variants and No-Indic, but remains just below the plain encoder. This is the main empirical point: source-specific conditional fusion is not merely a larger nonlinear head, it is most useful when external priors have different reliability profiles, and it still exposes where present auxiliary priors are incomplete.

Because many matched deltas are small, we compute paired bootstrap intervals over archived prediction files using 1000 resamples. Table~\ref{tab:bootstrap} is a conservative check on the two baseline families with complete archived prediction coverage. ToxGate has positive deltas in 17 of 20 settings against Plain and 15 of 20 against Concat, with positive 95\% intervals in 14 and 12 settings. The source-specific ablation conclusions are summarized separately in Table~\ref{tab:focused}.

\begin{table}[t]
\caption{Paired bootstrap summary over archived Plain/Concat/ToxGate prediction files. $\Delta$F1 is ToxGate minus comparison model. Positive/negative CI means the 95\% interval lies entirely above/below zero.}
\label{tab:bootstrap}
\centering
\small
\begin{tabular}{lccc}
\toprule
\textbf{Comparison} & \textbf{$\Delta$F1 $>$ 0} & \textbf{Positive CI} & \textbf{Negative CI} \\
\midrule
ToxGate vs. Plain & 17 / 20 & 14 / 20 & 2 / 20 \\
ToxGate vs. Concat & 15 / 20 & 12 / 20 & 1 / 20 \\
\bottomrule
\end{tabular}
\end{table}

\subsection{Robustness and Diagnostics}
We corrupt auxiliary sources at test time by shuffling, dropping, or perturbing Detoxify, Indic, rule, or all auxiliary features. This test separates useful dependence from brittle dependence. Concat changes very little under corruption, which is not strong evidence of robustness by itself. Together with its weaker average results, this suggests that static concatenation often learns to ignore the auxiliary scores. MLP is much more brittle, especially when Detoxify or all auxiliary features are removed. ToxGate sits between these extremes: it uses auxiliary evidence enough to improve transfer and severe slices, but its drops under corruption are bounded. Its gate means are narrow, usually around the middle of the sigmoid range, so we do not interpret gates as hard explanations. They are better read as soft residual modulation: a diagnostic of source reliance, not a faithful causal explanation of every decision.

\begin{table}[t]
\caption{Robustness under auxiliary corruption, averaged over datasets and encoders. Values are macro-F1 changes from the clean setting. Smaller drops are better, although near-zero drops can also mean that a model has learned to ignore the auxiliary features.}
\label{tab:interventions}
\centering
\small
\setlength{\tabcolsep}{5pt}
\begin{tabular}{lccc}
\toprule
\textbf{Intervention} & \textbf{Concat} & \textbf{MLP} & \textbf{ToxGate} \\
\midrule
Shuffle Detox & -0.000 & -0.097 & -0.027 \\
Shuffle Indic & -0.000 & -0.001 & -0.000 \\
Shuffle Rule & -0.000 & -0.002 & -0.000 \\
Noise all auxiliary & -0.000 & -0.016 & -0.001 \\
Drop all auxiliary & -0.000 & -0.149 & -0.024 \\
\bottomrule
\end{tabular}
\end{table}

Table~\ref{tab:reliability_map} gives a direct conditional-reliability map for the auxiliary sources on the two Hinglish transfer corpora. The slice sizes are English profanity $n=6{,}527$, Romanized slur $n=3{,}239$, violent threat $n=419$, and benign slang or quoted-abuse false alarms $n=199$. For the first three slices, entries report precision when the source fires followed by coverage of positive examples in the slice. The final column reports the false-trigger rate on non-abusive slang or quoted-abuse examples, where lower is better. Detoxify is reliable and high-coverage for English profanity, but fires on many benign slang examples and covers only a small fraction of Romanized slurs. The rule prior has the opposite profile: lower general coverage, but complete coverage on the violent-threat slice and the lowest benign-slang false-trigger rate.

\begin{table}[t]
\caption{Conditional reliability map for auxiliary priors. Entries are precision/coverage when a source fires, except the final false-trigger-rate column. Detoxify fires when any Detoxify score is at least 0.5, the Indic prior fires at 0.5, and the rule prior fires when the severity score is nonzero.}
\label{tab:reliability_map}
\centering
\small
\setlength{\tabcolsep}{4pt}
\resizebox{\textwidth}{!}{%
\begin{tabular}{lccccc}
\toprule
\textbf{Source} & \textbf{Overall precision} & \textbf{English profanity} & \textbf{Romanized slur} & \textbf{Violent threat} & \textbf{Benign slang FTR} \\
\midrule
Detoxify & 0.920 & 0.962 / 0.985 & 0.944 / 0.067 & 0.934 / 0.833 & 0.628 \\
Indic prior & 0.694 & 0.955 / 0.488 & 0.916 / 0.126 & 0.860 / 0.590 & 0.447 \\
Rule severity & 0.893 & 0.864 / 0.056 & 0.905 / 0.544 & 0.845 / 1.000 & \textbf{0.055} \\
\bottomrule
\end{tabular}
}
\end{table}


Table~\ref{tab:qualitative-gates} gives representative diagnostic examples with
auxiliary scores and learned source gates. Since each gate is vector-valued, we
report the mean activation of the corresponding gate vector. These values should
be read as diagnostics of source reliance, not as faithful causal explanations of
individual predictions.

\begin{table}[t]
\centering
\caption{Qualitative diagnostics for auxiliary scores and learned source gates.
Post excerpts are sanitized. Scores and gates lie in ([0,1]). Gate values are
mean activations over the corresponding source gate vector.}
\label{tab:qualitative-gates}
\scriptsize
\setlength{\tabcolsep}{3pt}
\begin{tabular}{p{2.3cm}p{2.7cm}cccccccc}
\toprule
Context & Sanitized excerpt & Gold & Pred. &
($t_{\mathrm{detox}}$) & ($t_{\mathrm{indic}}$) & ($t_{\mathrm{rule}}$) &
($g_{\mathrm{detox}}$) & ($g_{\mathrm{indic}}$) & ($g_{\mathrm{rule}}$) \\
\midrule
English profanity/slur
& \texttt{i ant all the b***hes}
& 1 & 0.66 & 0.06 & 0.83 & 0.00 & 0.50 & 0.47 & 0.46 \\

Romanized Hindi slur
& \texttt{Teri Man ki ch** Modi bahan ke lo**e kutte madr**hod}
& 1 & 1.00 & 0.16 & 0.69 & 0.00 & 0.47 & 0.47 & 0.46 \\

Violent threat
& \texttt{m** jao chalo waisay population bht zaida ho gi ha..}
& 1 & 0.64 & 0.01 & 0.25 & 0.00 & 0.47 & 0.47 & 0.46 \\

Benign slang / quoted abuse
& \texttt{@username ... Tumhari jiji ne hi khud bola tha mai [kamini] hu}
& 0 & 0.52 & 0.00 & 0.27 & 0.00 & 0.47 & 0.47 & 0.46 \\
\bottomrule
\end{tabular}
\end{table}

Qualitative mining supports the same reading. Detoxify is most useful for English profanity, the Indic cue helps on some Romanized slurs, and the rule cue helps violent or sexual threats. False positives arise when English toxicity priors overreact to slang or quoted abuse, and false negatives remain when abuse is implicit or requires social context. The practical diagnostic is therefore not ``the gate explains the label,'' but ``external priors are useful only when the linguistic context makes them trustworthy.''

\section{Moderation Triage}
The deployment-oriented value of ToxGate is decision support: routing high-confidence abusive posts to moderation review, uncertain or conflicting cases to human review, and low-risk posts to no action. ToxGate is designed as a triage aid, not an autonomous enforcement system, and any production use requires human oversight and disparate-error-rate evaluation. We simulate this by sorting test examples by predicted abuse probability and evaluating the top 10\% high-risk bucket. Table~\ref{tab:focused} reports the high-risk precision comparison against fusion ablations, while Table~\ref{tab:triage} reports detailed Plain/ToxGate decision-support metrics. Across archived settings, ToxGate improves high-risk precision from 0.930 to 0.945 and high-risk macro-F1 from 0.479 to 0.485, in transfer settings, precision improves from 0.846 to 0.883. This is preliminary, but it aligns with the paper's central message: the benefit is concentrated in high-risk review regions, not uniformly across all examples.

\begin{figure}[t]
\centering
\scriptsize
\resizebox{\textwidth}{!}{%
\begin{tikzpicture}[
    box/.style={draw, align=center, minimum height=8mm, minimum width=21mm, inner sep=3pt},
    route/.style={draw, align=center, minimum height=7mm, minimum width=27mm, inner sep=3pt},
    arrow/.style={-{Latex[length=2mm]}, thick},
    node distance=8mm
]
\node[box] (input) {Input\\post};
\node[box, right=of input] (encoder) {Text\\encoder};
\node[box, above=of encoder] (priors) {External\\priors};
\node[box, right=of encoder] (gate) {Reliability\\gate};
\node[box, right=of gate] (score) {Moderation\\score};
\node[route, right=of score, yshift=9mm] (high) {High-confidence\\abusive};
\node[route, right=of score] (uncertain) {Uncertain or\\conflicting};
\node[route, right=of score, yshift=-9mm] (low) {Low-risk};
\draw[arrow] (input) -- (encoder);
\draw[arrow] (encoder) -- (gate);
\draw[arrow] (priors) -- (gate);
\draw[arrow] (gate) -- (score);
\draw[arrow] (score) -- (high);
\draw[arrow] (score) -- (uncertain);
\draw[arrow] (score) -- (low);
\end{tikzpicture}
}
\caption{Deployment-oriented interpretation of trust-aware fusion. The figure is a decision-support framing, not a claim of production readiness or autonomous enforcement.}
\label{fig:pipeline}
\end{figure}
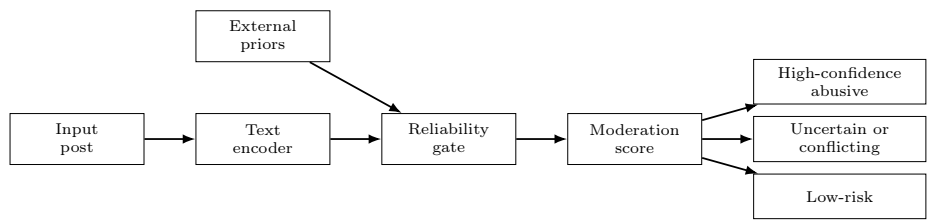

\begin{table}[t]
\caption{Moderation triage simulation over archived prediction files. The high-risk bucket is the top 10\% of examples by predicted abuse probability, averaged over five seeds and the relevant dataset--encoder settings.}
\label{tab:triage}
\centering
\small
\setlength{\tabcolsep}{4pt}
\begin{tabular}{lcccc}
\toprule
\textbf{Setting} & \textbf{Model} & \textbf{High-risk precision} & \textbf{High-risk F1} & \textbf{Uncertain rate} \\
\midrule
All settings & Plain & 0.930 & 0.479 & 0.070 \\
All settings & ToxGate & \textbf{0.945} & \textbf{0.485} & \textbf{0.062} \\
Transfer & Plain & 0.846 & 0.454 & 0.135 \\
Transfer & ToxGate & \textbf{0.883} & \textbf{0.467} & \textbf{0.114} \\
\bottomrule
\end{tabular}
\end{table}

\section{Limitations and Ethics}
The study has five main limitations. First, it uses three binary-label datasets, so it does not cover all languages, scripts, target groups, or fine-grained abuse taxonomies. Second, the Indic auxiliary score comes from a public external checkpoint, we freeze it and never use it to define labels or splits, but we cannot fully audit its upstream training data for overlap. Third, the focused ablation rows are aggregate diagnostics, per-dataset confidence intervals for SharedGate, ScalarGate, and source-filtered variants would further sharpen the causal decomposition. Fourth, five seeds on one fixed split measure optimization variance, not full resampling uncertainty; paired bootstrap intervals over predictions help but do not replace repeated splits. Fifth, slice labels and rule heuristics are operational diagnostics rather than sociolinguistic analysis.

Automated abuse detection also raises ethical risks. Toxicity tools can encode biases against dialects, marginalized communities, reclaimed language, or identity terms~\cite{borkan2019nuanced}. Trust-aware fusion can reduce reliance on unreliable signals in some cases, but it does not eliminate bias. Any practical use should keep humans in the loop, expose uncertainty, and evaluate disparate error rates before deployment.

\section{Conclusion}
This paper demonstrate Indian code-mixed moderation as a problem of conditional reliability. External toxicity tools and priors are useful, but they should not be trusted blindly under code-mixing, transliteration, slang, and language mismatch. ToxGate provides a simple source-aware mechanism for learning when to use English toxicity, Indic abuse, and rule-based severity priors. The evidence is strongest in the focused diagnostics: matched gains over plain encoders, transfer between Hinglish datasets, severe-abuse slices, and high-risk triage precision. The final lesson is practical: moderation systems should treat toxicity priors as conditional evidence, not as fixed features or ground truth.

\bibliographystyle{splncs04}
\bibliography{custom}

\end{document}